\pdfoutput=1
\documentclass[11pt]{article}

\usepackage[]{acl}

\usepackage{times}
\usepackage{latexsym}

\usepackage[T1]{fontenc}
\usepackage[utf8]{inputenc}

\usepackage{microtype}
\usepackage{graphicx}
\usepackage{amssymb,amsmath,latexsym,amsfonts,amsthm,cleveref,bm,bbm}
\usepackage{booktabs}
\usepackage{algorithm}
\usepackage{algorithmic}
\usepackage{multirow}

\title{Visual Information Guided Zero-Shot Paraphrase Generation}

\author{First Author \\
  Affiliation / Address line 1 \\
  Affiliation / Address line 2 \\
  Affiliation / Address line 3 \\
  \texttt{email@domain} \\\And
  Second Author \\
  Affiliation / Address line 1 \\
  Affiliation / Address line 2 \\
  Affiliation / Address line 3 \\
  \texttt{email@domain} \\}
\author{Zhe Lin\and Xiaojun Wan \\
 Wangxuan Institute of Computer Technology, Peking University \\
 Center for Data Science, Peking University \\
 The MOE Key Laboratory of Computational Linguistics, Peking University \\
 {\tt \{linzhe,wanxiaojun\}@pku.edu.cn} \\}

\begin{document}
\maketitle
\begin{abstract}
% There are several works leverage image caption dataset to train paraphrasing model by regarding different captions of the same image as paraphrase. This may lead to a huge semantic shift as different captions may describe different elements in the image. In this paper, we propose \textbf{v}isual \textbf{i}nformation guided zero-shot \textbf{p}araphrase \textbf{g}eneration (ViPG) which leverage image-caption pairs to train paraphrasing model. We adopt the multi-modal joint encoder to encode the image feature and text feature and joint train image captioning model and paraphrasing model. During training, we employ cross entropy loss to train image captioning model and leverage symmetric KL divergence to align the image feature and the text feature. In inference, we only send text to our model to generate paraphrase. Both automatic evaluation and human evaluation show our model can generate paraphrase with excellent relevancy, fluency and diversity.

Zero-shot paraphrase generation has drawn much attention as the large-scale high-quality paraphrase corpus is limited. Back-translation, also known as the pivot-based method, is typical to this end. Several works leverage different information as ``pivot'' such as language, semantic representation and so on. In this paper, we explore using visual information such as image as the ``pivot'' of back-translation. Different with the pipeline back-translation method, we propose \textbf{v}isual \textbf{i}nformation guided zero-shot \textbf{p}araphrase \textbf{g}eneration (\textbf{ViPG}) based only on paired image-caption data. It jointly trains an image captioning model and a paraphrasing model and leverage the image captioning model to guide the training of the paraphrasing model. Both automatic evaluation and human evaluation show our model can generate paraphrase with good relevancy, fluency and diversity, and image is a promising kind of pivot for zero-shot paraphrase generation. 

\end{abstract}

\section{Introduction}

Paraphrase generation is a long-standing problem for natural language processing that aims to rewrite a text in other forms while preserving its original semantics. Paraphrase generation has many applications in other down-stream tasks, such as machine translation \cite{mehdizadeh-seraj-etal-2015-improving}, semantic parsing \cite{berant-liang-2014-semantic} and so on.

With the development of supervised seq2seq generation, most paraphrase systems depend on the large-scale aligned paraphrase corpora to train the seq2seq model. This leads to the fact that the quality of aligned corpora is extremely important for training a paraphrase system. However, high-quality paraphrase datasets are still lacking in many domains. To solve this problem, there are a few works focusing on zero-shot paraphrase generation such as back-translation. Back-translation makes use of language as pivot and treats the back-translated text as the paraphrase of the original text. For example, \citet{mallinson-etal-2017-paraphrasing} leveraged multilingual neural machine translation to generate paraphrase and \citet{cai-etal-2021-revisiting} proposed to employ semantic representation as the ``pivot language'' of back-translation to generate paraphrase. All these works show that back-translation can generate high-quality paraphrase.

% There are many ways to create paraphrase datasets, such as back-translation, synonyms on Wikipedia and so on.
% A lot of works also leverage image caption datasets to train paraphrasing models by creating aligned data from different captions of the same image \cite{gupta2017deep,cao-wan-2020-divgan,lin-wan-2021-pushing}. These works hold the assumption that different captions of an image can be seen as paraphrases, and leverage these caption pairs to train their paraphrasing models. However, different captions of an image may describe different elements in the image, an example is shown in figure \ref{cap-pair}, which leads to huge semantic differences between different captions. These caption pairs can not be regarded as valid aligned paraphrase datas. Training paraphrasing model with image caption datasets is an urgent problem to be explored.

\begin{figure}
\centering
\includegraphics[scale=0.74]{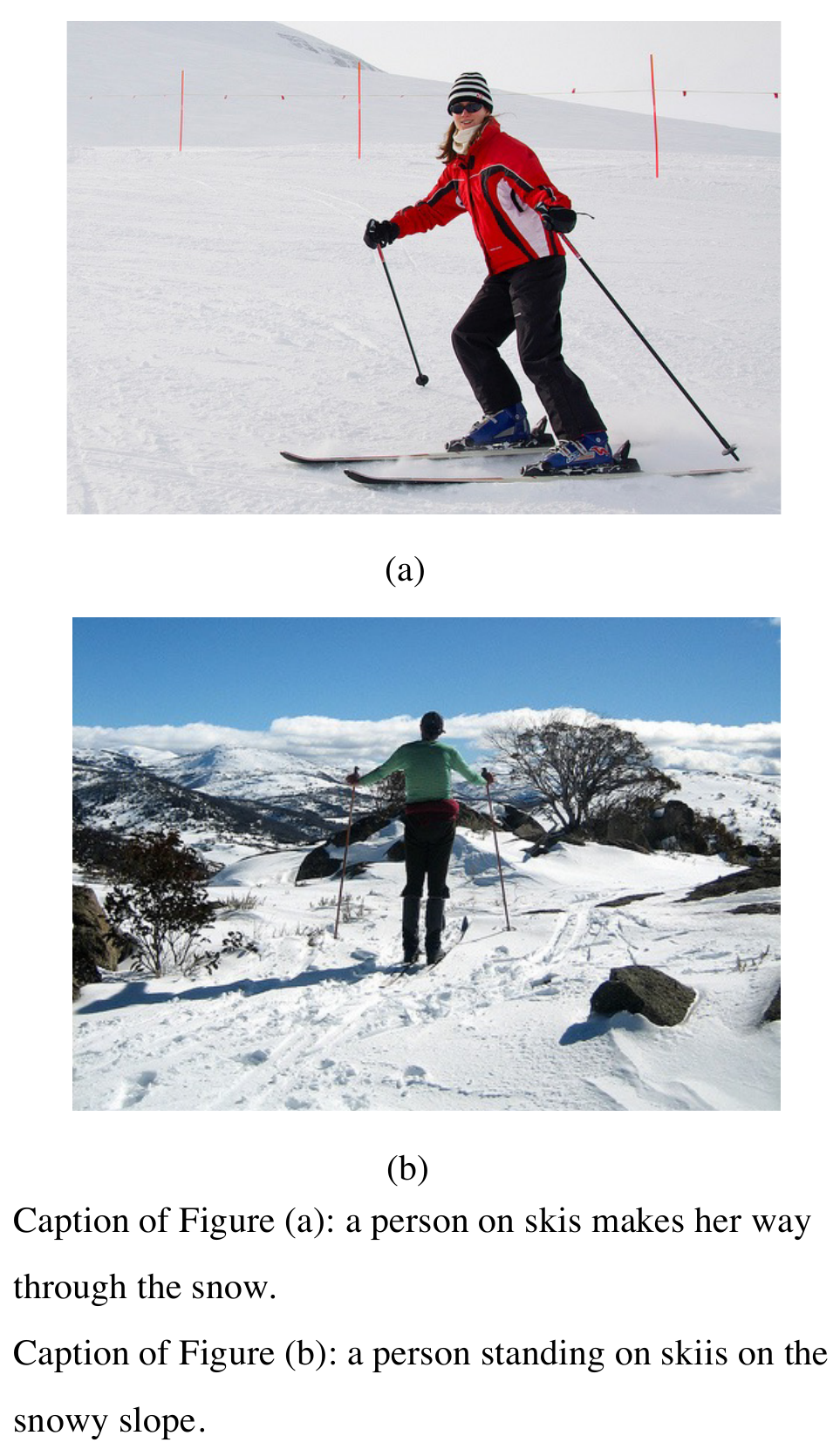}
\caption{An example that similar images may have different captions.}
\label{cap}
\end{figure}

Inspired by back-translation based paraphrase generation, we explore to leverage visual information (i.e. image in this study) to guide the zero-shot paraphrase generation as similar images or similar partial images may have different captions or descriptions that can be treated as paraphrases. Figure \ref{cap} shows an example. A naive method is that we can use image as the ``pivot language'' and generate paraphrase by back-translation with a text-to-image model and an image-to-text model. Unfortunately, text-to-image generation is still a challenging task and it is hard to generate an image of sufficient quality from the text. The semantic loss in text-to-image generation is so huge that generating paraphrases using this method is barely possible. Another method is that we can use an image captioning model to generate a caption from the image corresponding to the original text, and regard this caption as the paraphrase of the original text to train a supervised paraphrasing model. However, the generated caption may describe different elements in the image with the original text, which leads to huge semantic shift. 
% \textbf{Question: did you do experiments to compare with another method?}

In this study, we propose \textbf{v}isual \textbf{i}nformation guided zero-shot \textbf{p}araphrase \textbf{g}eneration (\textbf{ViPG}), which leverages image information to guide the paraphrase generation based only on paired image-caption data. 
We jointly train a specific image captioning model and a paraphrasing model, and leverage the output of the image captioning model to guide the training of the paraphrasing model. This can be regarded as distilling the knowledge of the image captioning model to the paraphrasing model at the word level. 

% Instead of using the pipeline back-translation method, we train the image captioning model and leverage image caption to guide the training of the paraphrasing model in real-time. 

% employ a shared-parameter model to train the image caption and paraphrase generation together. For image caption, we optimize the cross-entropy loss between the generated caption and the ground truth. For paraphrase generation, we optimize the symmetric KL divergence between the output of the image captioning model and paraphrasing model to align the text feature with the image feature. This can be regarded as distilling the knowledge of the image captioning model to the paraphrasing model at the word level. During inference, we only send the text to the model to generate its paraphrase.

Experiment results on two datasets show our model substantially outperforms the supervised paraphrasing model trained on paired caption-caption data and it can generate valid paraphrases with high diversity. We also compare our model with other zero-shot paraphrase generation methods such as autoencoder and back-translation, and analyze the strengths and weaknesses of these methods.  In all, our contributions can be summarized as follows:

\begin{itemize}
    \item To the best of our knowledge, we are the first to explore to leverage visual information to guide zero-shot paraphrase generation.
    
    \item We propose a novel model to leverage visual information to guide zero-shot paraphrase generation. Our method jointly trains an image captioning model and a paraphrasing model, and employs this image captioning model to guide the training of the paraphrasing model.  Our code is publicly available at \url{https://github.com/L-Zhe/ViPG}.
    
    \item Empirical studies on two image caption datasets show the effectiveness of our model and the image is demonstrated to be a promising kind of pivot for zero-shot paraphrase generation.
\end{itemize}

\section{Related Works}

There are several works leveraging image caption datasets like MSCOCO to train the paraphrasing model. \citet{prakash-etal-2016-neural} proposed residual-LSTM model to generate paraphrase. \citet{gupta2017deep} found deep generative model such as variational auto-encoder can achieve better performance in paraphrase generation. \citet{fu2019paraphrase} regraded the bag of word as the latent variable of VAE to control the semantic of paraphrase. \citet{chen-etal-2020-semantically} proposed a semantically consistent and syntactically variational encoder-decoder framework, which uses adversarial learning to ensure the syntactic latent variable be semantic-free. \citet{cao-wan-2020-divgan} leverage GAN to generate multiple diverse paraphrases. \citet{lin-wan-2021-pushing} raised multi-round paraphrase generation to improve the diversity and leveraged back-translation to maintain the semantic.  All these works regard different captions of the same image as paraphrase and leverage caption-caption pairs to train paraphrasing model.

There are also some works focus on zero-shot paraphrase generation. \citet{mallinson-etal-2017-paraphrasing} revisited back-translation paraphrase generation with neural machine translation. \citet{cai-etal-2021-revisiting} leveraged AMR as the new pivot of back translation. \citet{thompson-post-2020-paraphrase} proposed a novel decoding strategy to generate diverse paraphrase via multilingual translation. \citet{liu-etal-2020-unsupervised} leveraged simulated annealing to train unsupervised paraphrase generation model.

% \begin{figure*}
% \centering
% \includegraphics[scale=0.6]{img/overview_big.pdf}
% \caption{The overview architecture of our proposed model, which includes a multi-modal joint encoder and a parameter-sharing decoder. $v_{\mathcal{I}}$, $v_{\mathcal{O}}$, $v_{\mathcal{R}}$ are the tag vectors that indicate the different types of input.}
% \label{overview}
% \end{figure*}

\begin{figure}[htp]
\centering
\includegraphics[scale=0.51]{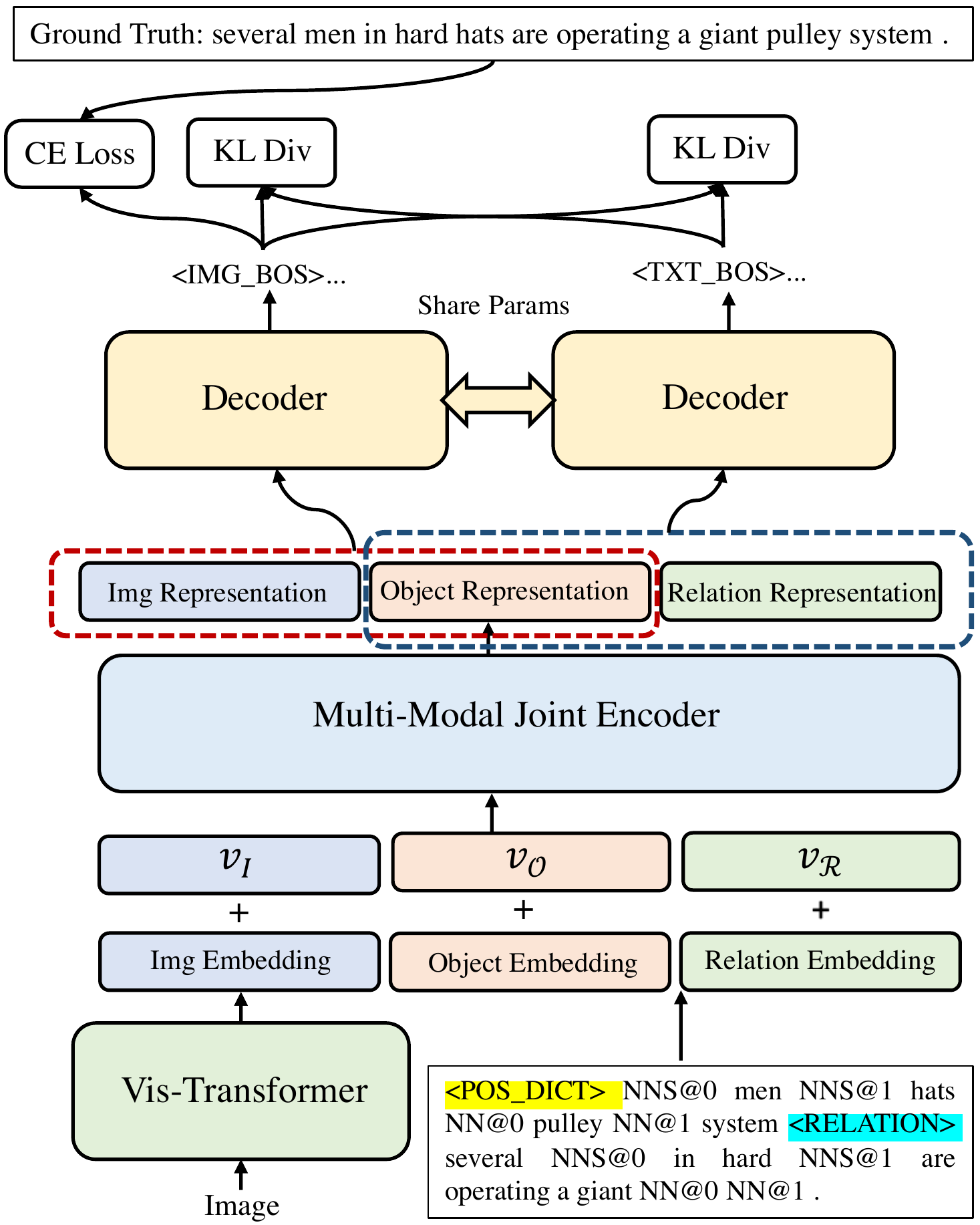}
\caption{The overview architecture of our proposed model, which includes a multi-modal joint encoder and a parameter-sharing decoder. $v_{\mathcal{I}}$, $v_{\mathcal{O}}$, $v_{\mathcal{R}}$ are the tag vectors that indicate the different types of input.}
\label{overview}
\end{figure}

\section{Methodology}

As mentioned earlier, the pipeline back-translation with the ``image pivot'' can not generate valid paraphrase as the performance of the text-to-image generation model is poor. Different from the pipeline method, our proposed method jointly trains an image captioning model and a paraphrasing model, and leverages the output of the image captioning model to guide the training of the paraphrasing model. The rationale is that an image may correspond to different captions with same meaning\footnote{Note that the different captions provided by human judges for a same image in most existing datasets like MSCOCO are often semantically inconsistent, so we do not aim to make use of the caption pairs to train the paraphrase model in this study.}. The image captioning model with our specific design (i.e., with additional input of object representations) may generate a caption that is different from the original caption for an input image while keeping the same meaning, and this output caption can be treated as the paraphrase of the original caption and it can be used for training the paraphrasing model. Our model relies only on image captioning dataset consisting of pairs of image and caption, and it does not need any text paraphrasing corpus and any data of caption pairs of same image. 
Each pair in the training dataset includes an image $\mathcal{I}$ and a corresponding caption sentence $\mathcal{S}=\{w^1,\cdots,w^N\}$, where $w^i$ is the $i$-th word of the sentence and $N$ is the sentence length.

In this section, we begin by introducing the initial embeddings of the image and text, followed by describing our multi-modal joint encoder, which employs partial attention to encode the image and the text together. Then we introduce a decoder with masked object copy mechanism to guide text generation. Finally, the objective functions will be detailed. The overall architecture of our model is shown in Figure \ref{overview}.

\subsection{Initial Image and Text Embeddings}

\subsubsection{Image Embedding}

For an input image $\mathcal{I}$, we first leverage Vision Transformer \cite{dosovitskiy2021an} to encode the image into an embedding matrix $\tilde{E}_{\mathcal{I}}$ as its excellent performance in many vision tasks. We further use $v_{\mathcal{I}}$ as a tag embedding vector to indicate the image tag. After that, the initial image representation $E_{\mathcal{I}}$ is obtained as follows:

\begin{equation}
\begin{aligned}
    \tilde{E}_{\mathcal{I}} &= \operatorname{ViT}(\mathcal{I}) \\
    E_{\mathcal{I}} &= \tilde{E}_{\mathcal{I}} + v_{\mathcal{I}}
\end{aligned}
\end{equation}

\noindent where $\operatorname{ViT}$ is the Vision Transformer encoder, $v_{\mathcal{I}} \in \mathbb{R}^d$ is the learnable parameter and $E_{\mathcal{I}} \in \mathbb{R}^{l \times d}$, where $d$ is the feature's dimension and $l$ is the patch length split by Vision Transformer. $+$ operation between a matrix and a vector means that the vector is added to all components of the matrix at the dimension of sequential length.

Note that we use the Vision Transformer to get $\tilde{E}_{\mathcal{I}}$ and fix it during the training of our model. This can save a bunch of training resources and has been proved to be reliable in many multi-modal tasks.

\subsubsection{Text Embedding}

A caption sentence can only describe the main elements of an image rather than all the details, and existing image captioning model tends to generate different captions talking about different objects for an image, which may cause semantic shift when using such image captioning model to guide the paraphrasing model. To tackle this problem, we extract the object words from the caption sentence and use them to help the image captioning model to generate more accurate and consistent captions.

Specifically, we regard nouns in a sentence as objects and the rest part of the sentence as the relation of these objects. We create the object sequence for all nouns in the sentence in this format: {\ttfamily\{POS\_TAG@index WORD\}}, where {\ttfamily POS\_TAG} is the part-of-speech of this word, {\ttfamily index} is used to distinguish different words of the same {\ttfamily POS\_TAG}. We replace all nouns in the sentence with their corresponding {\ttfamily POS\_TAG@index}. We regard the processed sequence as the relation described by the sentence. Then we concatenate the object sequence and the relation sequence as the input text. Table \ref{txt_split} shows an example of the transformed input text.

\begin{table}[htb]
\centering
\small
\scalebox{1}{
\begin{tabular}{p{7.1cm}}
\toprule[1.2pt]
\boxed{\textbf{Original Text:}} several men in hard hats are operating a giant pulley system . \cr
\boxed{\textbf{Object Sequence:}} NNS@0 men NNS@1 hats NN@0 pulley NN@1 system \cr
\boxed{\textbf{Relation Sequence:}} several NNS@0 in hard NNS@1 are operating a giant NN@0 NN@1 . \cr
\boxed{\textbf{Transformed Input Text:}} {\ttfamily <POS\_DICT>} NNS@0 men NNS@1 hats NN@0 pulley NN@1 system {\ttfamily <RELATION>} several NNS@0 in hard NNS@1 are operating a giant NN@0 NN@1 . \cr
\bottomrule[1.2pt]
\end{tabular}}
\caption{An example about splitting a text to the object sequence and relation sequence.}
\label{txt_split}
\end{table}

We denote the embedding matrices of the object sequence and relation sequence as $\tilde{E}_{\mathcal{O}}$ and $\tilde{E}_{\mathcal{R}}$, respectively. We also add the embedding matrices with different tag embedding vectors to indicate different parts of the input information (i.e., object or relation). Finally, we combine these two parts of information as a whole and add positional encoding.

\begin{equation}
\begin{aligned}
    \hat{E}_{\mathcal{O}} &= \tilde{E}_{\mathcal{O}} + v_{\mathcal{O}} \\
    \hat{E}_{\mathcal{R}} &= \tilde{E}_{\mathcal{R}} + v_{\mathcal{R}} \\
    [E_{\mathcal{O}}, E_{\mathcal{R}}] & = [\hat{E}_{\mathcal{O}}, \hat{E}_{\mathcal{R}}] + W_{PE}
\end{aligned}
\end{equation}

\noindent where $v_{\mathcal{O}}, v_{\mathcal{R}} \in \mathbb{R}^d$ are learnable parameters, $W_{PE}$ is the positional encoding matrix, $\left[*,*\right]$ is concatenation operation at the dimension of sequential length.

\subsection{Multi-Modal Joint Encoder}

We adopt Transformer encoder architecture as multi-modal joint encoder to further encode the image and text. In order to reduce the gap between image representation and text representation, we share the encoder parameters instead of leveraging separate encoders for image and text. We concatenate the initial image embedding $E_{\mathcal{I}}$ with the initial text embedding $[E_{\mathcal{O}}, E_{\mathcal{R}}]$ and send them to the encoder at the same time.

% \begin{figure}[htb]
% \centering
% \includegraphics[scale=0.36]{img/attn.pdf}
% \caption{The architecture of partial attention.}
% \label{attn}
% \end{figure}

The powerful performance of the Transformer encoder is due to its self-attention structure, as each element in the sequence can aggregate the whole sequential information with dynamic attention weight. However, this global attention is not suitable for our model as our image captioning model and paraphrasing model should focus on different input information. Instead, we just want the image feature to focus on the information from itself and the object feature. While the image information should be ignored when encoding the text feature. Based on the rules above, we introduce the partial attention as follows:

\begin{equation}
\small
\begin{aligned}
    \tilde{I}_i &= \operatorname{MHAttn}(I_{i-1}, [I_{i-1}, O_{i-1}], [I_{i-1}, O_{i-1}]) \\
    \tilde{O}_{i} &= \operatorname{MHAttn}(O_{i-1}, O_{i-1}, O_{i-1}) \\
    \tilde{R}_{i} &= \operatorname{MHAttn}(R_{i-1}, [O_{i-1}, R_{i-1}], [O_{i-1}, R_{i-1}]) \\
    % \hat{I}_i &= \operatorname{LayerNorm}(\tilde{I}_i + I_{i-1}) \\
    % \hat{O}_i &= \operatorname{LayerNorm}(\tilde{O}_i + O_{i-1}) \\
    % \hat{R}_i &= \operatorname{LayerNorm}(\tilde{R}_i + R_{i-1}) 
\end{aligned}
\end{equation}

\noindent where $\operatorname{MHAttn}(Q, K, V)$ is the multi-head attention \cite{vaswani2017attention}, $I_{i-1},O_{i-1},R_{i-1}$ are the learned representation matrices of the image, object sequence and relation sequence at the $(i-1)$-th layer. $\tilde{I}_i$, $\tilde{O}_i$, and $\tilde{R}_i$ are then fed into FFN module followed by residual connection and layer normalization that are the same as the vanilla Transformer encoder to get the representation matrices at the  $i$-th layer. We employ $I,O,R$ to represent the encoding representations of the image, object sequence and relation sequence at the last layer respectively.

\subsection{Decoder}

Decoder aims to generate text from the encoding feature. Different from the text sent into the encoder which is split into the object sequence and the relation sequence, the decoder directly generates the original text $\mathcal{S}$.

Our model includes a caption decoder and a paraphrase decoder. The caption decoder generates the caption from the image feature representation and the object feature representation, and the paraphrase decoder generates the paraphrase corresponding to the original text. We share the parameters of these two decoders, and leverage different {\ttfamily BOS} tokens to guide the decoder to deal with different features. We leverage {\ttfamily <IMG\_BOS>} to guide the caption generation and employ {\ttfamily <TXT\_BOS>} to guide the paraphrase generation. The details of the decoder are as follows:

\begin{equation}
\begin{aligned}
    D_{\mathcal{I}} &= \operatorname{Decoder}([I, O],\ \mathtt{<IMG\_BOS>}) \\
    D_{\mathcal{S}} &= \operatorname{Decoder}([O, R],\ \mathtt{<TXT\_BOS>}) \\
    \tilde{P}_{\mathcal{I}} &= \operatorname{softmax}(W_oD_{\mathcal{I}} + b_o) \\
    \tilde{P}_{\mathcal{S}} &= \operatorname{softmax}(W_oD_{\mathcal{S}} + b_o)
\end{aligned}
\end{equation}

\noindent where $\operatorname{Decoder}(\mathtt{feature}, \mathtt{BOS token})$ is the Transformer decoder, $W_o, b_o$ are learnable parameters, which map the dimension of output features $D_{\mathcal{I}}$, and $D_{\mathcal{S}}\in \mathcal{R}^{N \times d}$ to the size of vocabulary.

We add copy mechanism \cite{see-etal-2017-get} to guide the decoder to generate the correct objects. We only copy from the object sequence rather than the whole sentence. The copy probabilities are calculated as follows:

\begin{equation}
\begin{aligned}
    P^c_{\mathcal{I}} &= \operatorname{softmax}(D_{\mathcal{I}}^{\top}O)\\
    P^c_{\mathcal{S}} &= \operatorname{softmax}(D_{\mathcal{S}}^{\top}O)\\
\end{aligned}
\end{equation}

Copy mechanism can improve the semantic accuracy of the generated text but may lead to low diversity of the object words. There may be more than one way to describe an object, and copying the object words from the original sentence directly can lose this diversity. Therefore we employ the masked object copy mechanism to avoid excessive copy. We randomly mask 20\% object words in the object sequence as {\ttfamily <UNK>} during the copy process. This can help the model learn to generate the diverse object words rather than copy from the original sentence directly. The final output probabilities of the image caption and the paraphrase generation are denoted as $P_{\mathcal{I}}=\{p_\mathcal{I}^1,\cdots,p_\mathcal{I}^N\}$ and $P_{\mathcal{S}}=\{p_\mathcal{S}^1,\cdots,p_\mathcal{S}^N\}$, respectively.

\subsection{Loss Function}

We employ cross-entropy loss to train the image captioning model as follows:

\begin{equation}
\begin{aligned}
    \mathcal{L}_{ce} = -\frac{1}{N}\sum_{i=1}^N\operatorname{log}p^i_{\mathcal{I}}(w^i)
\end{aligned}
\end{equation}

\noindent where $p^i_{\mathcal{I}}(w^i)$ is the corresponding probability of $w^i$ in $p^i_{\mathcal{I}}$.

For the paraphrasing model, we do not directly optimize the cross-entropy loss based on $P_{\mathcal{S}}$ as this can lead to the degeneration of the model into an autoencoder. On the contrary, we align the information from the two models by reducing the gap between $P_{\mathcal{I}}$ and $P_{\mathcal{S}}$. Inspired by the R-Drop \cite{liang2021rdrop}, we optimize the symmetric KL divergence between $P_{\mathcal{I}}$ and $P_{\mathcal{S}}$ as follows:

\begin{equation}
\begin{aligned}
    \mathcal{L}_{kl} = -\frac{1}{2N}\sum_{i=1}^N\{\operatorname{KL}(p^i_{\mathcal{I}} || p^i_{\mathcal{S}}) + \operatorname{KL}(p^i_{\mathcal{S}} || p^i_{\mathcal{I}}) \}
\end{aligned}
\end{equation}

We train image captioning and paraphrase generation together and the total loss of our model is as follows:

\begin{equation}
\begin{aligned}
    \mathcal{L} = \mathcal{L}_{ce} + \lambda \mathcal{L}_{kl}
\end{aligned}
\end{equation}

\noindent where $\lambda$ is a hyper-parameter.

\subsection{Inference}

Although we leverage image-caption pair to train our model, the image is not required during the inference. In the inference phase, we split the original text to the object sequence and the relation sequence and leverage {\ttfamily <TXT\_BOS>} to guide the paraphrase generation.

\section{Evaluation Setup}

\subsection{Datasets}

Two image caption corpora (\textbf{MSCOCO}\footnote{\url{https://cocodataset.org}} and \textbf{Flickr30k}\footnote{\url{https://shannon.cs.illinois.edu/DenotationGraph}}) are used as our evaluation datasets. The MSCOCO dataset includes $118,287$ images and Flickr30k includes $31,783$ images, each image in both dataset has five different captions. We construct two types of training datasets for each corpus:  1) One-caption: We randomly sample one caption for each image and thus only one image-caption pair per image is used for training; 2) All-captions: We use all five captions to create five image-caption pairs per image. 
For each dataset, we randomly sample $4000$ captions as validation dataset. For MSCOCO, we leverage all $2,5014$ captions provided by the official validation dataset for test. For Flickr30k, we randomly sample $8000$ captions as the test dataset. Note that there is no ground-truth paraphrase for each caption in the validation and test datasets and we do not need them in our evaluation at all\footnote{We do not use datasets like Parabank and Quora for evaluation because these datasets are in totally different domains with our training datasets, and thus we use the in-domain caption data for evaluation in this study.}.  

% We also employ \textbf{Google Conceptual Captions}\footnote{\url{https://ai.google.com/research/ConceptualCaptions/download}} dataset for data augmentation. We select the data with the caption's length greater than 10, and there are about 880K image-caption pairs. We combine these pairs with the pairs in the one-caption dataset from MSCOCO and Flickr30k to train our model.

\subsection{Competitive Methods}

The competitive methods used for comparison are mainly in three categories:

\textbf{Supervised models trained with caption-caption pairs:} Following previous works, we regard different captions of an image as paraphrases and leverage these caption-caption pairs to train a Transformer model as the supervised paraphrase generation model. And we also finetune the Bart model \cite{lewis-etal-2020-bart} with the caption-caption pairs. Besides, we take one caption in the caption-caption pair as the ``reference'' paraphrase of the other caption and evaluate the ``reference'' paraphrase as well.

\textbf{AutoEncoder models with diversity decoding strategies:} We train the Transformer and Bart models as the AutoEncoder models respectively. For both models, we leverage various decoding strategies including greedy search, top-k decoding and top-p decoding to generate diverse paraphrases.

\textbf{Pipeline back-translation methods with various kinds of pivot:} We employ language, AMR graph and image as pivots separately. For back-translation with language pivot, we leverage English-German translation systems provided by \citet{DBLP:journals/corr/abs-1907-06616}. For back-translation with AMR pivot, we generate paraphrase according to \citet{cai-etal-2021-revisiting}. For back-translation with image pivot, we leverage text-to-image model provided by \citet{DBLP:journals/corr/abs-2107-02423} to generate image from text and leverage image captioning model provided by \citet{rennie2017self} to generate its correspond caption as the paraphrase of the original text.
\begin{table*}
    \centering
    % \small
    \scalebox{0.87}{
    \begin{tabular}{l|ccc|ccc}
    \toprule[1.2pt]
    \multirow{2}{1in}{\textbf{Model}} & \multicolumn{3}{c|}{\textbf{MSCOCO}} & \multicolumn{3}{c}{\textbf{Flickr30k}} \cr
    &\textbf{Self-BLEU}$\downarrow$&\textbf{BERTScore}$\uparrow$&\textbf{PPL}&\textbf{Self-BLEU}$\downarrow$&\textbf{BERTScore}$\uparrow$&\textbf{PPL} \cr
    \hline
    Source                      & -     & -     & 178.82 & -     & -     & 234.11 \cr
    \hline
    \multicolumn{3}{l}{\textbf{Supervised models trained with caption-caption pairs:}}&&&& \cr
    Caption Reference           & 8.01  & 49.83 & 177.55 & 7.02  & 47.62 & 195.37 \cr
    Transformer                 & 14.81 & 57.30 & 116.96 & 13.00 & 56.31 & 363.15 \cr
    Bart(Fine Tune)           & 19.61 & 61.29 & 85.15  & 19.46 & 62.33 & 278.37 \cr
    \hline
    \multicolumn{3}{l}{\textbf{AutoEncoder models with diversity decoding strategies:}}&&&& \cr
    Bart(Original)              & 99.89 & 99.94 & 178.24 & 99.91 & 99.97 & 233.94 \cr
    \quad + top-k(k=5)          & 99.82 & 99.90 & 177.61 & 99.74 & 99.89 & 233.16 \cr
    \quad + top-p(p=0.9)        & 99.86 & 99.92 & 177.97 & 99.85 & 99.94 & 233.70 \cr
    AutoEncoder                 & 92.19 & 95.16 & 213.15 & 85.54 & 90.85 & 309.26 \cr
    \quad + top-k(k=5)          & 84.30 & 90.55 & 284.53 & 74.17 & 83.88 & 428.08 \cr
    \quad + top-p(p=0.9)        & 74.69 & 82.60 & 530.60 & 62.56 & 72.74 & 815.54 \cr
    \hline
    \multicolumn{3}{l}{\textbf{Pipeline back-translation methods:}}&&&& \cr
    BackTranslation-AMR         & 36.63 & 75.51 & 353.13 & 32.63 & 75.52 & 430.10 \cr
    BackTranslation-Language    & 54.17 & 84.17 & 202.05 & 53.87 & 84.75 & 258.28 \cr
    BackTranslation-Image       & 9.06  & 51.22 & 104.79 & 4.55  & 45.26 & 81.78 \cr
    \hline
    \textbf{ViPG(Ours):}&&&&&& \cr
    One-Caption                 & 38.25 & 71.95 & 130.24 & 29.12 & 66.11 & 159.06 \cr
    All-Captions                & 43.40 & 76.38 & 155.61 & 31.21 & 69.54 & 359.66 \cr
    
    \bottomrule[1.2pt]
    \end{tabular}}
    \caption{\label{result} Automatic evaluation results. The evaluation metrics include diversity, semantic relevancy and fluency.}
\end{table*}
\subsection{Metrics}

We evaluate our model in three aspects: diversity, relevancy and fluency. We leverage \textbf{Self-BLEU}, which calculates the BLEU score between the paraphrase and the original sentence, to evaluate the diversity of the paraphrase. We leverage \textbf{BERTScore} to measure the semantic relevancy. For fluency, we employ GPT-Large without finetuning to calculate the perplexity scores (\textbf{PPL}) of different models' outputs.

In addition, we perform human evaluation of model outputs with respect to diversity, relevancy and fluency. All ratings were obtained using a five point Likert scale. We randomly sample 200 instances, including 100 from MSCOCO and 100 from Flickr30k. We employ 6 graduate students to rate each instance, and we ensure every instance is rated by at least three judges.

\subsection{Training Details}

We leverage Vision Transformer base\footnote{The Vision Transformer model we used is available at \url{https://huggingface.co/google/vit-base-patch16-224-in21k}} to generate the initial image embedding with the dimension of 768. In order to align image features, we also set the latitude of encoder and decoder to 768. We set $\lambda$ to 1 in loss function. Other hyper-parameters are same to the vanilla Transformer. We select the model with highest BERTScore on the validation dataset. During inference, we leverage beam search with 5 beam size to generate paraphrase.

\section{Results}

\begin{table}
    \centering
    % \small
    \scalebox{0.9}{
    \begin{tabular}{lc|cc|cc}
    \toprule[1.2pt]
    \multicolumn{2}{c|}{\multirow{2}{*}{\textbf{Model}}} & \multirow{2}{*}{\textbf{Rel.}} &  \multirow{2}{*}{\textbf{Flu.}} &\multicolumn{2}{c}{\textbf{Div.}} \cr
    % \cline{5-6}
    & & & & \textbf{Lexi.} & \textbf{Synt.} \cr
    \hline
    \multicolumn{2}{l|}{Caption Reference}              & 2.36 & 3.46 & 3.16 & 2.80 \cr
    \multicolumn{2}{l|}{Transformer}                    & 2.81 & 3.40 & 3.31 & 3.03 \cr
    \multicolumn{2}{l|}{Bart(fine tune)}                & 2.28 & 3.89 & 3.47 & 3.11 \cr
    \multicolumn{2}{l|}{AutoEncoder(top-k)}             & 4.28 & 2.39 & 2.37 & 2.20 \cr
    \multicolumn{2}{l|}{BT-Language}                    & 3.91 & 3.51 & 3.43 & 3.40 \cr
    \multicolumn{2}{l|}{BT-AMR}                         & 3.54 & 3.39 & 3.20 & 3.88 \cr
    \multicolumn{2}{l|}{BT-Image}                       & 1.39 & 3.09 & 2.73 & 2.59 \cr
    \hline
    \multicolumn{2}{l|}{ViPG(One-Caption)}              & 3.78 & 3.72 & 3.71 & 3.42 \cr
    \multicolumn{2}{l|}{ViPG(All-Captions)}              & 3.73 & 3.64 & 3.60 & 3.34 \cr
    \bottomrule[1.2pt]
    \end{tabular}}
    \caption{Human evaluation results. BT means BackTranslation. Rel., Flu. and Div. is the abbreviation of relevancy, fluency and diversity. Lexi. and Synt. mean lexical and syntactic, respectively.}
    \label{human}
\end{table}

% \begin{table*}
%     \centering
%     \small
%     \scalebox{0.9}{
%     \begin{tabular}{l|cccc|cccc}
%     \toprule[1.2pt]
%     \multirow{2}{1in}{\textbf{Dataset}}& \multicolumn{4}{c|}{\textbf{MSCOCO}} & \multicolumn{4}{c}{\textbf{Flickr30k}} \cr
%     &\textbf{SIZE}&\textbf{Self-BLEU}$\downarrow$&\textbf{BERTScore}$\uparrow$&\textbf{PPL}&\textbf{SIZE}&\textbf{Self-BLEU}$\downarrow$&\textbf{BERTScore}$\uparrow$&\textbf{PPL} \cr
%     \hline
%     One caption           & 150K & 38.25 & 71.95 & 130.24 & 150K & 29.12 & 66.11 & 159.06 \cr
%     Multiple captions         & 590K & 43.40 & 76.38 & 155.61 & 16K & 31.21 & 69.54 & 359.66 \cr
%     Google Conceptual Caption       & 1M & 40.82 & 73.65 & 115.03 & 1M& 31.12 & 67.71 & 120.03 \cr
%     \bottomrule[1.2pt]
%     \end{tabular}}
    
%     \caption{\label{ablation} Automatic evaluation results for using different training datasets.}
% \end{table*}

\begin{table*}[htp]
    \centering
    % \small
    \scalebox{0.86}{
    \begin{tabular}{|p{4cm}|p{12cm}|}
        \hline
        \multicolumn{2}{|c|}{\textbf{Cases from MSCOCO}}\\
        \hline
         Original & a cup , toothbrushes , and other items sit on the side of a small sink .\\
        \hline
         Transformer(supervised) &	a bathroom sink and its reflection in the mirror .\\
        \hline
         Bart(Fine tune) &  a bathroom sink with toothbrushes and other bathroom items .\\
        \hline
         AutoEncoder(top-k) & a cup , toothbrushes , and other items sit on the side of a small sink . \\
        \hline
         AutoEncoder(top-p) & a cup , toothbrushes , and other items sit on the side of a small sink . \\
        \hline
         BackTranslation-Language & a cup , toothbrushes and other objects lie on the side of a small sink . \\
        \hline
         BackTranslation-AMR & cups , toothbrushes and other items are sat on the side of the small sink .\\
         \hline
         BackTranslation-Image&  a toothbrush sitting on top of a sink .\\
         \hline
         ViPG(One-Caption) &  a cup contains toothbrushes and other items on the side of a sink .\\
         \hline
         ViPG(All-Captions) &  a cup filled with toothbrushes and other items sitting on the side of a sink .\\
         \hline
         \multicolumn{2}{c}{}\\
         \hline
         \multicolumn{2}{|c|}{\textbf{Cases from Flickr30k}}\\
        \hline
         Original & a woman and child stand on the beach while sailboats sail on the ocean .\\
        \hline
         Transformer(supervised) & a mom and son enjoying the beach .\\
        \hline
         Bart(Fine tune) &  a woman and child are standing on a beach by sailboats .\\
        \hline
         AutoEncoder(top-k) & a woman and child stand on the beach while sailboats sail on the ocean . \\
        \hline
         AutoEncoder(top-p) & a woman and child stand on the fattening beach while the ocean sail glances bieber strussel tugs installment on swatch transports palomitas the woman . \\
        \hline
         BackTranslation-Language & a woman and child stand on the beach while sailboats sail the ocean . \\
        \hline
         BackTranslation-AMR & women and children stand on the beach as boats sail in the ocean .\\
         \hline
         BackTranslation-Image&  two people walking on the beach with a boat .\\
         \hline
         ViPG(One-Caption) &  a woman and a child on the beach with sailboats in the ocean .\\
         \hline
         ViPG(All-Captions) &  a woman and child are on the beach looking at sailboats in the ocean .\\
         \hline
    \end{tabular}}
    \caption{Examples from MSCOCO and Flickr30k and the generated paraphrases by different models.}
    \label{casestudy}
\end{table*}

\subsection{Result Analysis}

Tables \ref{result} and \ref{human} show the results of automatic evaluation and human evaluation, respectively.

For supervised models trained with caption-caption pairs, the big semantic gap between the outputs of these models and the original sentence can be obvious from the low BERTScore. There are also great semantic differences between the caption reference and the original sentence. Using paired caption-caption data to train the paraphrasing model can lead to a huge semantic shift. The result of human evaluation also shows that supervised models trained with caption-caption pairs may generate paraphrase that changes the semantic of the original sentence, which can not be regarded as valid paraphrase.

For AutoEncoder models, they all get the high BERTScore but high self-BLEU, which means that the paraphrase generated by these models lacks diversity. Since Bart is a pretrained autoencoder model, top-k and top-p decoding strategies can hardly introduce diversity. For AutoEncoder,  the diversity decoding strategy can indeed increase the paraphrase diversity, and yet it is harmful to the fluency of the generated sentence. The diversity decoding strategy can lead to a significant increase in PPL, this means that the quality of the generated paraphrase is affected. The human evaluation also shows the decline of sentence fluency caused by the diversity decoding strategy.

For pipeline back-translation methods, BackTranslation-AMR and BackTranslation-Language can generate good paraphrase with enough relevancy and diversity. From the human evaluation, we find that the paraphrase generated by BackTranslation-AMR has stronger diversity than BackTranslation-Language. BackTranslation-AMR can introduce diversity at syntactic level as the AMR is an abstract semantic representation of a sentence. However, BackTranslation-Image can not generate valid paraphrase with adequate semantic relevancy, this is because text-to-image generation is still a challenge task and may cause a huge semantic shift. In case study, we also show an example of BackTranslation-Image for a more intuitive explanation.

For our ViPG model, we solve the semantic shift in BackTranslation-Image and get the adequate BERTScore. Beside, our model performs well on diversity and fluency. Our model gets the low self-BLEU which means high diversity. For fluency, our model also achieves the best PPL score among all valid paraphrasing models. The human evaluation shows that the diversity of our model is mainly at lexical level, while syntactic diversity also performs well. Briefly, our model performs much better than other paraphrasing models leveraging image-caption data and has strong competitiveness with zero-shot paraphrasing models.

We also find that the BERTScore has a significant improvement for our ViPG model trained by all-captions dataset, but the human evaluation scores of diversity and fluency have decreased. This means that using all captions of an image to create training dataset is harmful for our model. 

% \subsection{Ablation Study}

% As an image has multiple captions for most image caption datasets, we perform ablation study on different ways of constructing training dataset. Concretely, we employ two methods: an image and a correspond caption to be an image-caption pair and an image and multiple captions to be different image-caption pairs.
% From the table we can find that leverage all captions to create dataset
% may lead the decline of fluency though this method can create a bigger training dataset. Although BERTScore has improved significantly for multiple captions dataset, the human evaluation of diversity and fluency have decreased. This means that the dataset created from all captions of the same is harmful for our model. 

% We also employ Google Conceptual Captions dataset for data enhancement. The result shows that increate the volume of training data create from an image and a corresponding caption can indeed improve the quality of paraphrase.

\subsection{Ablation Study}

  We perform the ablation study on MSCOCO to investigate the influence of different modules in our ViPG model. We replace the transformed input text with the original text to explore the effect of embedding nouns and relations separately. We remove the $\operatorname{KL}(p^i_{\mathcal{I}} || p^i_{\mathcal{S}})$ and $\operatorname{KL}(p^i_{\mathcal{S}} || p^i_{\mathcal{I}})$ separately to show the influence of symmetric KL divergence in loss function. To further explore the effect of the masked object copy mechanism, we conduct another two experiments. One of the experiments we remove the masked object copy mechanism. In another experiment, the copy mechanism can copy the words from the whole sentence, not just the object words. Table \ref{ablation} shows the results of the ablation study.
 
 \begin{table}
    \centering
    % \small
    \scalebox{0.9}{
    \begin{tabular}{l|ccc}
    \toprule[1.2pt]
    \textbf{Model} &\textbf{Self-B}$\downarrow$&\textbf{BS}$\uparrow$&\textbf{PPL} \cr
    \hline
    Origin & 38.25 & 71.95 & 130.24 \cr
    \hline
    Original Text & 33.28 & 53.55 & 438.73 \cr
    w/o $\operatorname{KL}(p^i_{\mathcal{I}} || p^i_{\mathcal{S}})$ & 7.89 & 12.44 & 530.98 \cr
    w/o $\operatorname{KL}(p^i_{\mathcal{S}} || p^i_{\mathcal{I}})$ & 28.90 & 35.48& 470.22 \cr
    w/o Copy Mechanism & 35.79 & 63.55 & 230.48\cr
    Copy the Whole Sent & 61.49 & 77.50 & 203.61\cr
        
    \bottomrule[1.2pt]
    \end{tabular}}
    \caption{\label{ablation} Self-B and BS is the abbreviation of self-BLEU and BERTScore.}
\end{table}

We can see that each module in our model does contribute to the overall performance. Using the original text directly can lead to significant degradation of BERTScore. The reason of huge semantic shift is that there are many objects in an image and the image-caption model can not distinguish which object is described in the original text. The two-part of symmetric KL divergence is necessary for the model training. The object copy mechanism can improve the relevancy and fluency of paraphrasing. However, copying the whole sentence without restriction can lead to a lack of diversity.

\subsection{Case Analysis}

We perform case studies for better understanding the model performance. Table \ref{casestudy} shows running examples from MSCOCO and Flickr30k. Obviously, there are some degrees of semantic shift for the paraphrases generated by supervised models such as Transformer and Bart. BackTranslation-Image generates paraphrases with high semantic loss. Our ViPG model can generate paraphrases with good diversity, relevancy and fluency. However, as the shortage of image caption dataset, the paraphrases generated by our model may introduce additional semantic information, such as the ``contains'' and ``filled with'' in the example from MSCOCO. This does not affect the readability of paraphrase, but still a problem to be solved.

\section{Conclusion}

In this paper, we propose a visual information guided zero-shot paraphrase generation approach. We explore employing image as the ``pivot'' of the back-translation. Instead of using a pipeline back-translation, we jointly train an image captioning model and a paraphrasing model together. We leverage the image captioning model to guide the training of the paraphrasing model. Both automatic evaluation and human evaluation show the competitive performance of our model. In the future, we will explore huge-scale image caption dataset to train our model and test the model's ability on other domains. Moreover, leveraging video as pivot for paraphrase generation is also an interesting research direction. 

\section*{Acknowledgments}

This work was supported by National Key R\&D Program of China (2021YFF0901502), National Science Foundation of China (No. 62161160339), State Key Laboratory of Media Convergence Production Technology and Systems and Key Laboratory of Science, Technology and Standard in Press Industry (Key Laboratory of Intelligent Press Media Technology). We appreciate the anonymous reviewers for their helpful comments. Xiaojun Wan is the corresponding author.

\bibliography{anthology,custom}
\bibliographystyle{acl_natbib}

\appendix

% \section{Example Appendix}
% \label{sec:appendix}

% This is an appendix.

\end{document}